\pdfoutput=1
\documentclass[11pt]{article}

\usepackage[]{acl}
\usepackage[most]{tcolorbox}
\usepackage{times}
\usepackage{latexsym}
\usepackage{graphicx}
\usepackage{booktabs}
\usepackage[normalem]{ulem}
\usepackage[export]{adjustbox}
\usepackage{amsmath}
\usepackage{breqn}
\usepackage{makecell}
\usepackage{tabularx}
\usepackage[T1]{fontenc}

\usepackage[utf8]{inputenc}

\usepackage{microtype}

\usepackage{enumitem}
\usepackage{subcaption}
\title{\textsc{CLICKER}: A Computational LInguistics Classification Scheme for Educational Resources}

\author{Swapnil Hingmire$^1$\thanks{\;\;Equal Contribution}, Irene Li$^3$\footnotemark[1], Rina Kawamura$^3$, Benjamin Chen$^3$, Alexander Fabbri$^3$\\
 \textbf{Xiangru Tang$^3$, Yixin Liu$^3$, Thomas George$^2$, Tammy Liao$^3$, Wai Pan Wong$^3$} \\
 \textbf{Vanessa Yan$^3$, Richard Zhou$^3$, Girish K. Palshikar$^1$, Dragomir Radev$^3$} \\
$^1$Tata Consultancy Services Limited, India \\
\texttt{swapnil.hingmire@tcs.com} \\
$^2$University of Waterloo, Canada, $^3$Yale University, USA \\
\texttt{\{irene.li,rina.kawamura,dragomir.radev\}@yale.edu}

    }

\begin{document}
\maketitle
\begin{abstract}
A classification scheme of a scientific subject gives an overview of its body of knowledge. It can also be used to facilitate access to research articles and other materials related to the subject. For example, the ACM Computing Classification System (CCS) is used in the ACM Digital Library search interface and also for indexing computer science papers. We observed that a comprehensive classification system like CCS or Mathematics Subject Classification (MSC) does not exist for Computational Linguistics (CL) and Natural Language Processing (NLP). We propose a classification scheme -- CLICKER for CL/NLP based on the analysis of online lectures from 77 university courses on this subject. The currently proposed taxonomy includes 334 topics and focuses on educational aspects of CL/NLP; it is based primarily, but not exclusively, on lecture notes from NLP courses.
We discuss how such a taxonomy can help in various real-world applications, including tutoring platforms, resource retrieval, resource recommendation, prerequisite chain learning, and survey generation. 
\end{abstract}

\section{Introduction}
As the scientific literature and educational resources continue to grow beyond an individual's capacity to follow them, an indexing and classification scheme can play an important role in facilitating access to different stakeholders. Additionally, a classification scheme of an academic subject provides a cognitive map of the domain. For example, the current Mathematics Subject Classification (MSC)\footnote{\url{https://mathscinet.ams.org/mathscinet/msc/msc2020.html}}, Medical Subject Headings (MeSH)\footnote{\url{https://meshb.nlm.nih.gov/}}, Physics Subject Headings (PhySH)\footnote{\url{https://physh.aps.org/}} provide both a cognitive map and a body of knowledge in Mathematics, the Life Sciences, and Physics respectively. These taxonomies are currently used to classify scientific articles on these subjects. In the context of computing, the ACM Computing Classification System (CCS)\footnote{\url{https://dl.acm.org/ccs}} has served as the standard for classifying the computing literature since 1964~\cite{Bernard_major_updates}. CCS is also used in the ACM Digital Library (DL) to ``index content for subject-oriented searching; to find \textit{similar} documents; to create author expertise profiles; to identify strong research areas in Institutional Profiles; and to create the topical tag clouds found in aggregated SIG and conference views''~\cite{Bernard_major_updates}.

We observed that unlike MSC, PhySH, and ACM-CCS, a comprehensive classification scheme does not exist for Computational Linguistics or Natural Language Processing (CL/NLP) and this makes it difficult to search for educational materials on a specific topic and scientific articles in the ACL Anthology (AA)\footnote{\url{https://www.aclweb.org/anthology/}}. We observed that several standard subject classification schemes focus on `Language' and `Linguistics'\footnote{In this paper, we assume ``class'', ``category'', ``research area'' and ``term'' are similar and we use them interchangeably.}, but not on CL/NLP. For example, there are Class P in the Library of Congress classification, class 400 and 410 in Dewey Decimal Classification, and class P in Ranganathan's Colon Classification~\cite{satija2017colon}). ACM-CCS has a set of classes specific to CL/NLP in the following category: \textit{CCS$\rightarrow$Computing methodologies$\rightarrow$Artificial intelligence$\rightarrow$Natural language processing} but its size is small. These classification schemes are not comprehensive enough and cannot cover the topics in the ACL Anthology. For example, none of the classification schemes mention even key areas of NLP such as `Summarization', `Question-Answering', or `Sentiment Analysis'.

In this position paper, we highlight the need for a classification scheme for CL/NLP, which, just like ACM-CCS, can be used to classify and index educational and research materials on this subject. Based on this, we propose a classification scheme -- CLICKER for CL/NLP based on the analysis of online lectures from a number of university courses on this subject. The folloing sections include (1) overview of the process of the classification scheme creation, (2) basic statistics of the the classification scheme, and (3) a list of possible applications.


\section{Existing Taxonomies}
In this section, we survey various existing taxonomies, including one Academic NLP Taxonomy, several NLP taxonomies from the Web, CS Taxonomies and Non-CS Taxonomies.

\subsection{NLP Taxonomies in Academia}
The ACL Anthology (AA) is the open-source archive of the proceedings of all ACL sponsored conferences and journal articles~\cite{bird-etal-2008-acl,RadevLREC,gildea-etal-2018-acl}. It currently hosts more than 60,000 papers. AA has served as a valuable resource to characterize the work of the ACL community. For example,~\cite{hall-etal-2008-studying} used latent Dirichlet Allocation (LDA; \cite{DBLP:journals/jmlr/BleiNJ03}) based topic models and observed the shift in ideas in the field of CL/NLP. \cite{anderson-etal-2012-towards} analyzed the computational history of AA using LDA; however, their analysis is people-centric. In another paper, \cite{anderson-etal-2012-towards} group topics into high-level categories. However, it is hard to understand the process of their grouping. For example, `Dialog' and `Summarization' are grouped into a category named 'Linguistic Supervision', while `Speech Recognition' is considered as a part of `Government' category. \cite{schumann2015tracing} also analyze paradigm changes in AA, however they focus only on `Machine Translation'. \cite{jurgens-etal-2018-measuring} also explore the evolution of AA, but through citation frames and not from a classification perspective. \cite{10.1162/tacl_a_00254} categorize research articles related to neural models of NLP from different perspectives such as linguistic information, the challenge set for evaluation of neural networks, and methods for adversarial examples in NLP. However, this analysis is only focused on neural models of NLP. \cite{DBLP:journals/corr/abs-2107-12708} propose a taxonomy related to Question Answering (QA) and Reading Comprehension (RC) resources along multiple dimensions. The authors use the taxonomy to  categorize over 200 datasets related to QA/RC.  \cite{uban-etal-2021-studying} use topic modeling to track the evolution of topics in AA across three major dimensions: \textit{tasks, algorithms} and \textit{data}. 

\begin{table*}[t]
\small
\centering
\begin{tabular}{lllll}
\toprule
Name  & \#Concepts & Structure & Example Concepts \\ \midrule
NLPExplorer \cite{parmar2020nlpexplorer}         &   146    &   2    &     syntax, summarization, unsupervised             \\
The NLP Index        &   105    &   2    &    data augmentation, commonsense, bert     \\
nn4nlp-concepts     &   110    &   3   &    mini-batch SGD, transfer learning, transformer              \\ 
TutorialBank \cite{fabbri2018tutorialbank}        &   208    &  Prerequisites   &  em algorithm, attention models         \\ 
LectureBank \cite{li2019should}        &   322    &  Prerequisites   &  structured learning, named entity recognition         \\ 
NLP-progress  &  38    &  Flat   &  summarization, text classification         \\
ACM Computing Classification System     &   2113    &   6   &   serial architectures, information retrieval    \\
Computer Science Ontology   &   14164   &   multiple*   &   artificial intelligence, internet, bioindicator    \\
Semantic Scholar &  N/A & N/A &  Burst \\
Paper Reading  &    126   &    Flat   &   Bert, gradient descent, image classification    \\
\bottomrule
\end{tabular}
\caption{Existing Taxonomies. *We could not find the exact depth of the Computer Science Ontology. }
\label{tab:similar}
\end{table*}

\subsection{NLP Taxonomies from the Web}
Besides the ACL Anthology, other NLP Taxonomies exist. They have been developed by different groups of researchers, and they are not designed for the purpose of classifying publications, but rather for more general educational resources which are the same as our focus. 
We survey these similar NLP taxonomies in the following and compare them in Table \ref{tab:similar}. 


\textbf{NLPExplorer} \cite{parmar2020nlpexplorer} is an automatic portal for collecting, indexing, and searching CL and NLP papers from the ACL Anthology. The papers are indexed under a manually curated list of topics categorized into five broad categories, \emph{Linguistic Targets}, \emph{Tasks}, \emph{Approaches}, \emph{Languages}, and \emph{Dataset Types}. This approach is limited in that there are concepts that are not covered by the list of topics. Moreover, the two-level hierarchical structure does not allow for the further organization beyond the initial classification of topics.

\textbf{The NLP Index\footnote{\url{https://index.quantumstat.com/}}} is a search engine containing over 3,000 repositories of NLP-related code with their corresponding research papers. The repositories can be searched using dozens of pre-defined topic queries categorized into eight broad categories. Like NLPExplorer, this list of topics is not comprehensive and does not allow for the other categories beyond the initial ones.

\textbf{nn4nlp-concepts\footnote{\url{https://github.com/neulab/nn4nlp-concepts}}} is a concept hierarchy which attempts to cover the concepts needed to understand neural network models for NLP. The topics are generated both automatically and through manual annotation, and are organized in a hierarchical structure with a maximum depth of 3. However, these concepts are again only limited to neural network related topics.

\textbf{TutorialBank} \cite{fabbri2018tutorialbank} is a manually collected corpus of NLP educational resources, including research papers, blog posts, tutorials, abnd lecture slides. The most recent version\footnote{\url{https://aan.how/}} contains 23,193 resources. Together with the resources, they also propose a concept list containing 208 topics via crowdsourcing, with prerequisite annotations. Prerequisite relations are represented as a graph, instead of a tree structure for a taxonomy. 

\textbf{LectureBank} \cite{li2019should} is a manually collected dataset containing thousands of NLP-centric university-level lecture slides as well as 322 concepts collected through crowdsourcing. These concepts cover the field of NLP, basic machine learning, and deep learning for the purpose of prerequisite chain learning. However, like TutorialBank, these concepts have various granularity and forms in a graph structure.

\textbf{NLP-progress} is a repository to track NLP progress\footnote{\url{https://nlpprogress.com/}}. It contains the most advanced tasks and datasets, including entities from 14 languages. The current version contains 38 topics under English in a flat structure. 

\textbf{NLPedia}\footnote{\url{https://explainaboard.nlpedia.ai/leaderboard/}}  tracks the performance of more than 300 systems on 40 datasets and nine tasks. Additionally, they diagnose the strengths and weaknesses of a single system and interpret relationships between multiple systems \cite{DBLP:journals/corr/abs-2104-06387}.

\subsection{Computer Science Taxonomies}

The \textbf{ACM Computing Classification System\footnote{\url{https://dl.acm.org/ccs}}} is a poly-hierarchical classification scheme for the field of computing that can be utilized in semantic web applications. The 2,113 topics are organized in a tree structure with a maximum depth of six.

The \textbf{Computer Science Ontology\footnote{\url{http://cso.kmi.open.ac.uk/home}}} is a large-scale taxonomy of computer science research areas automatically generated from about 16 million publications. It includes 14,164 topics and 162,121 semantic relationships and is organized in a tree structure with "Computer Science" as the root. The ontology includes other semantic relationships such as equivalency between topics or indication.

\textbf{Semantic Scholar\footnote{\url{https://www.semanticscholar.org/}}} is an AI-powered research tool for scientific literature. The search engine covers research papers and filtering of results by field of study, date range, publication type, etc. The site includes literature in a large number of subjects, including computer science, and more specifically NLP. The papers are tagged with topic keywords that are automatically extracted using machine learning techniques. The site provides a Wikipedia summary for each topic and lists related, broader, and narrower topics, suggesting that these topics are organized in a hierarchical structure.

\textbf{ArnetMiner (AMiner)} \cite{tang2008arnetminer} is an online AI-powered service designed to perform search and data mining operations on academic publications. The service aims to build a social network of academic researcher profiles by identifying connections between researchers, conferences, and publications using graph techniques. This service covers a number of subjects, including Mathematics, Biology, and Forestry, in addition to Computer Science. 

\textbf{Papers with Code\footnote{\url{https://paperswithcode.com/}}} is an open resource with Machine Learning papers, code, datasets, methods and evaluation tables. The Machine Learning portal includes 255,497 papers with code, 2,217 methods, and 4,948 datasets. The methods are organized in a tree structure with seven top-level categories, which are further categorized into several levels. There are other portals for the different fields of Computer Science, Physics, Mathematics, Astronomy, and Statistics, although these portals have fewer resources. 

\textbf{Paper Reading\footnote{\url{http://paperreading.club/}}} is an index of research papers on Artificial Intelligence topics using pre-defined topic tags. In includes 126 topic tags but they are not organized in any particular structure.

\subsection{Non-CS Taxonomies}

Classification schemes exist for other subjects, e.g., the Mathematics Subject Classification taxonomy which covers mathematics topics, Medical Subject Headings (MeSH) for biomedical and health related topics, and Physics Subject Headings (PhySh) which covers Physics topics. These schemes are designed to help classify scientific literature on each of these respective subjects.

\section{Building CLICKER}

\subsection{Candidate Keyword Extraction}

The above mentioned existing taxonomies present fine-grained but limited topics in terms of the number and coverage.  To provide broader coverage of topics in CLICKER, we performed keyword extraction and analysis to generate a long candidate concept list which contains up to 10 thousand candidates. 

We believe that lecture notes contain clean and fine-grained topics, for example, the header of a page. We started with the current version of of LectureBank \cite{li2019should}, which includes more than 2,000 lecture slides and texts, converted to textual format.  We initially considered the lecture slide titles and the headers of each slide file page as a topic. We applied Textrank \cite{mihalcea2004textrank} to extract keywords and phrases. Finally, we ended up with a complete topic list of 4,397 candidates in descending order based on frequency.

Similarly, we extracted another candidate topic list from the TutorialBank \cite{fabbri2018tutorialbank} corpus. We only conducted keyword extraction on the resource titles to keep fine granularity. This resulted in a list consisting of 29,616 candidate topics. 

We then combined the different topics lists and sorted them by frequency. We kept the ones that appeared at least three times, which resulted in a clean list of 465 candidate topics.

\subsection{Additional Processing}

After some manual filtering, we ended up with our current version (1.0) of the taxonomy, which includes 271 topics that cover primarily CL and NLP topics, along with some related topics from Artificial Intelligence, Speech Processing, and Information Retrieval classes, as well as some prerequisite mathematical and statistical concepts. We skipped some spurious titles such as "Reading List" and "Midterm Information" and manually grouped some similar-sounding and overlapping topics. We show a set of sample slide titles in Table \ref{tab:slide_titles}.

\begin{table}[t]
\centering
\footnotesize
\begin{tabular}{ll}\toprule
\textbf{Most Frequent 15}   & \textbf{Less Frequent 15}                        \\ \midrule
recurrent neural networks     & relation extraction              \\
word embeddings               & conditional random fields        \\
neural networks               & adversarial search               \\
dependency parsing            & structured prediction models     \\
machine translation           & statistical machine translation  \\
language models               & probabilistic topic models       \\
logistic regression           & optimization \\
question answering            & computational discourse         \\
text classification           & lexicalized parsing              \\
convolutional neural networks & context free grammar             \\
hidden markov models          & bayesian networks                \\
distributional semantics      & expectation maximization         \\
language modeling             & kernel methods                   \\
lexical semantics             & compositional semantics          \\
support vector machines       & a* search and heuristics        \\ \bottomrule
\end{tabular}
\caption{Most frequent and less frequent topics extracted from slide titles. The right hand side include 15 of the less frequent topics, randomly selected.}
\label{tab:slide_titles}
\end{table}

\section{Applications}
In this section, we show how to use CLICKER in two interesting applications. The first one is to support resource classification in an existing educational platform; the second is to take advantage of the taxonomy to learn prerequisites between concepts in a transfer learning setting.

\subsection{CLICKER: Our Educational Platform}

\begin{figure}[t]
\centering
  \includegraphics[width=8cm,frame]{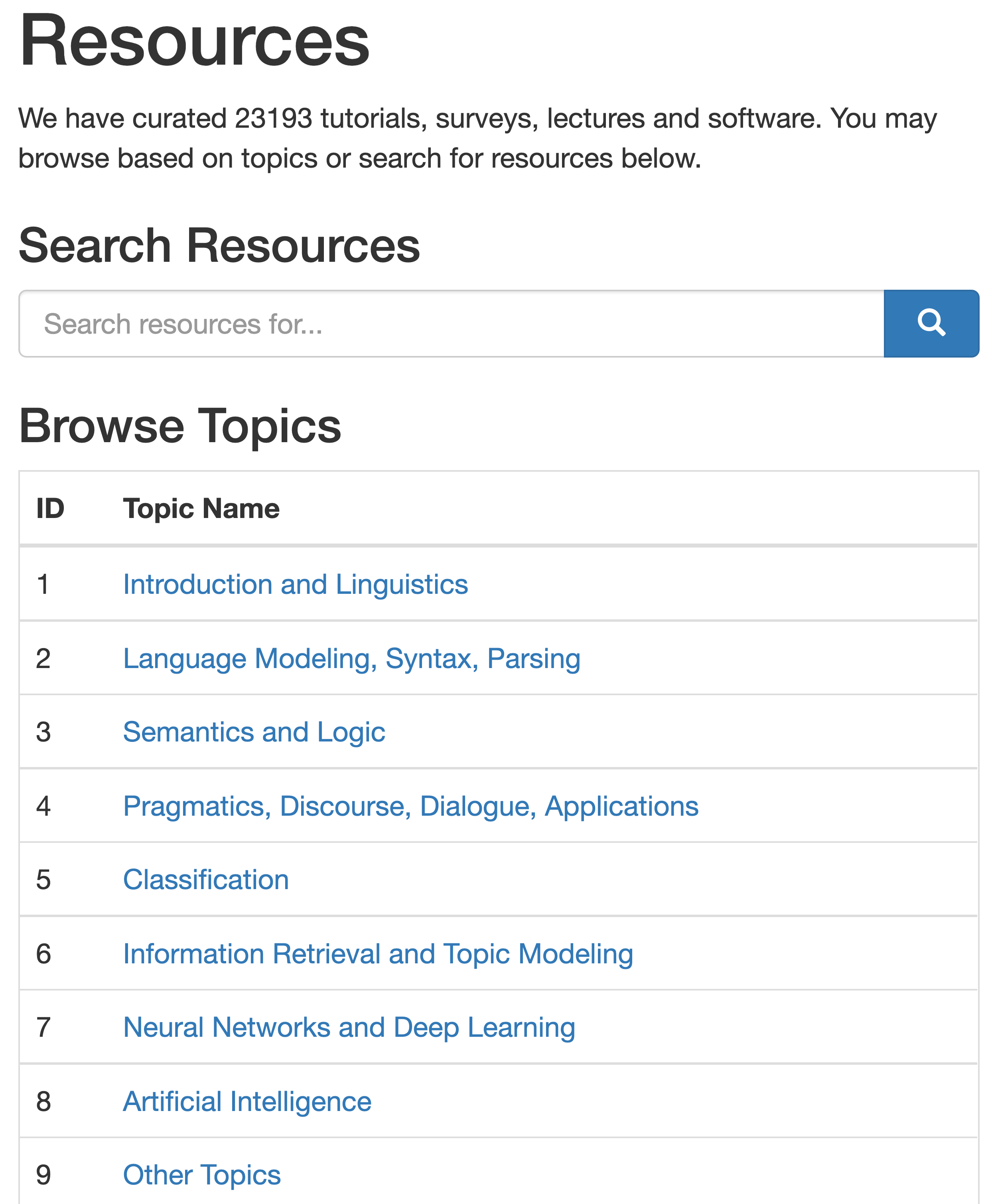}
  \caption{Top-level topics in AAN interface.}
  \label{fig:aan_top}
\end{figure}

\begin{figure}[t]
\centering
  \includegraphics[width=8cm,frame]{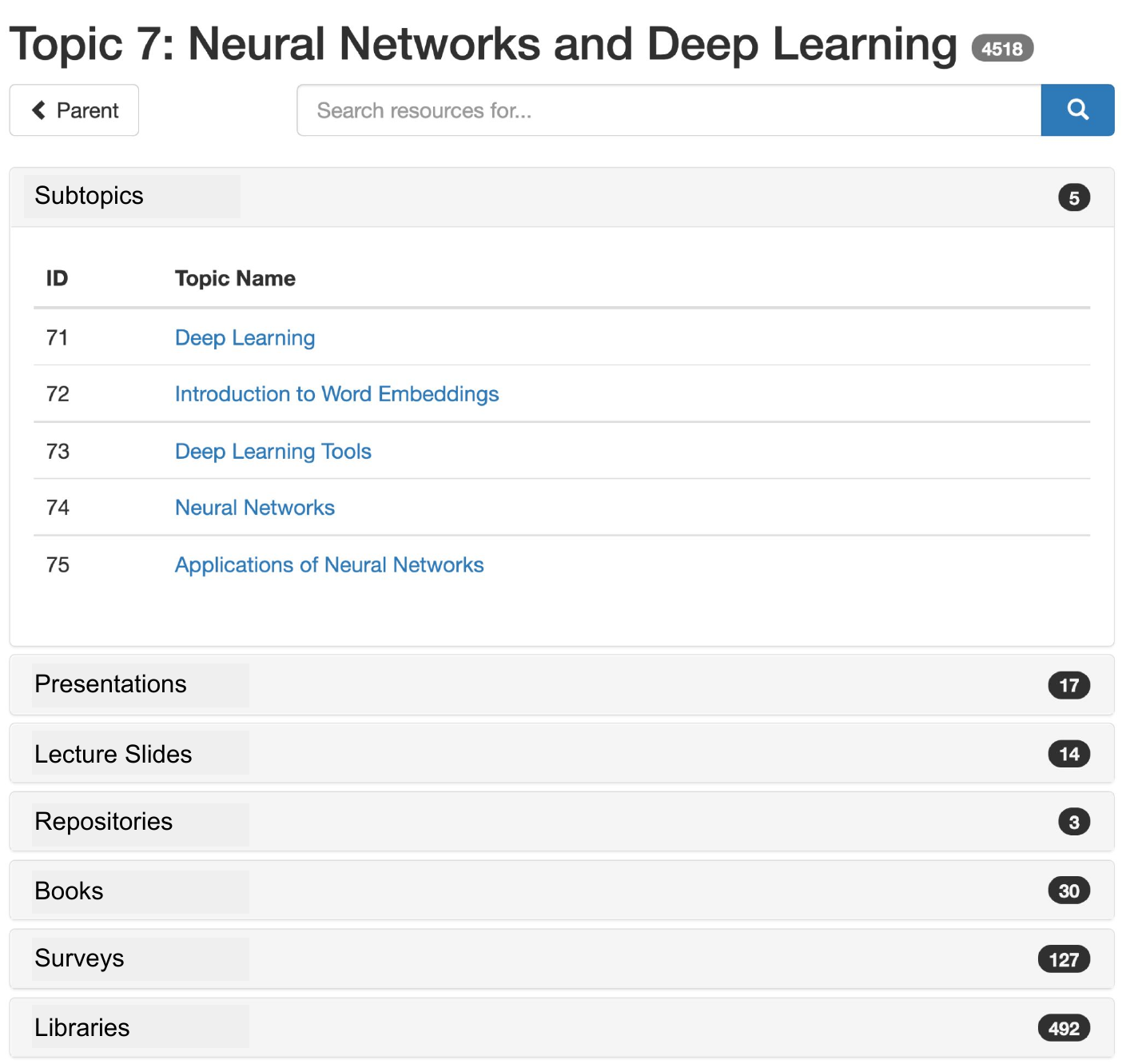}
  \caption{Second-level topics in AAN interface.}
  \label{fig:aan_sec}
\end{figure}

A taxonomy like CLICKER can be an important part of an online educational platform, AAN\footnote{\url{https://aan.how/}}. 
AAN encompasses a corpus of resources on NLP and related fields. These are educational resources primarily lectures and tutorials. A total of 23,293 resources have been manually assigned to nodes of the taxonomy. The website allows users to browse resources from the top-level of the taxonomy in the interface, as shown in Figure \ref{fig:aan_top}. Here we illustrate the nine top-level topics. Once the user clicks on a topic, they will see the next (finer) level of detail, as illustrated in Figure \ref{fig:aan_sec}.

\subsection{Resource Retrieval}

\begin{figure}[t]
\centering
  \includegraphics[width=8cm,frame]{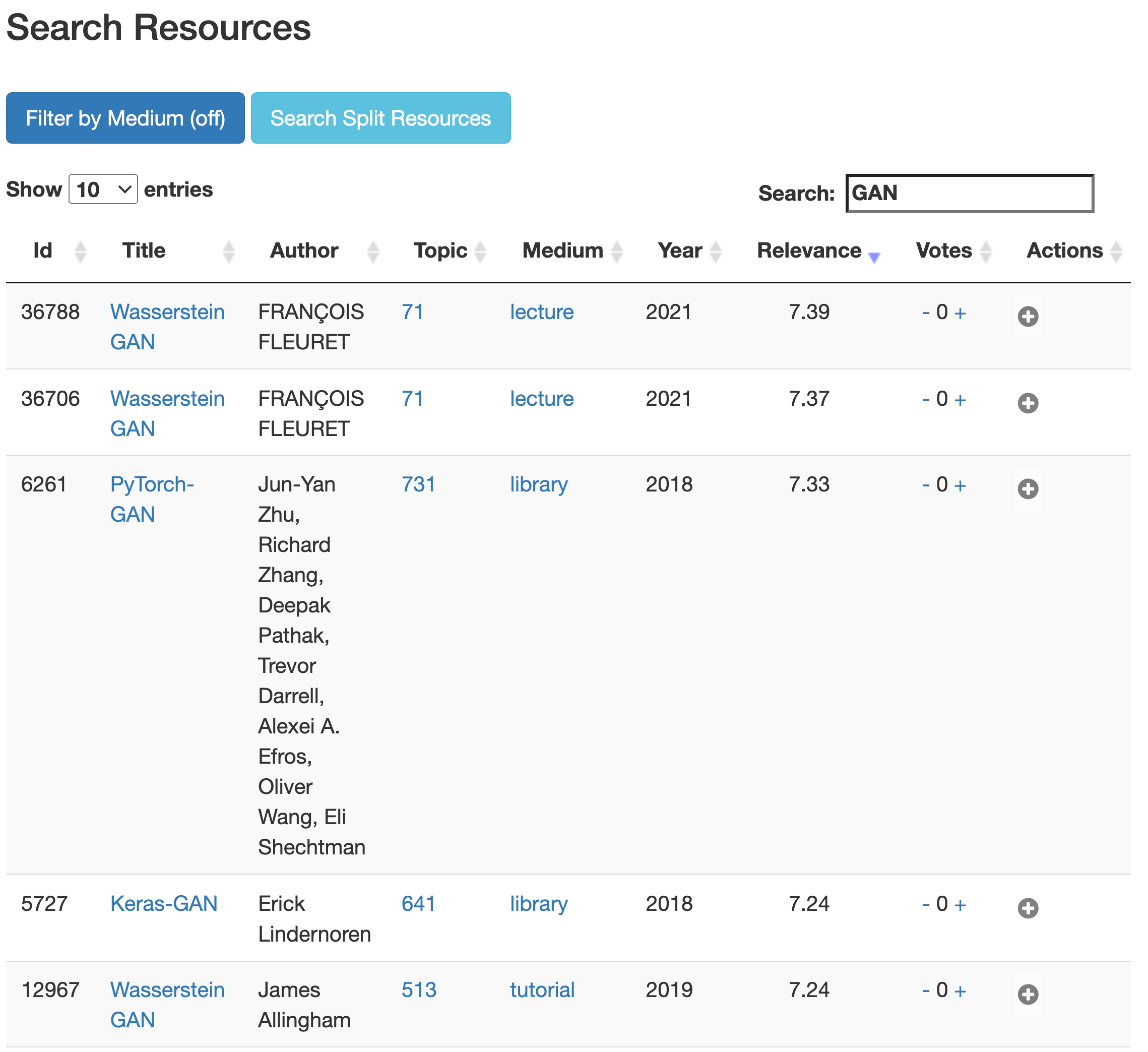}
  \caption{Resource retrieval results for the query \texttt{GAN} (the top 5 results). }
  \label{fig:aan_search}
  \vspace{-3mm}
\end{figure}

The educational platform can support several other applications. One of them is resource retrieval. Figure \ref{fig:aan_search} shows the top 5 results when a user types in the query keyword \texttt{GAN}, and the interface returns a list of resources based on relevancy. 
We utilized the Apache Lucene Core\footnote{\url{https://lucene.apache.org/core/}} as the text search engine. For each result, the interface also shows some metadata. The \textit{Topic} corresponds to the taxonomy node number. For example, the first resource belongs to (broad) topic 72 (Deep Learning) in Figure \ref{fig:aan_sec}. This interface supports keyword search as an alternative way of browsing resources mentioned in the previous section.

\subsection{Resource Recommendation}

\begin{figure}[t]
\centering
\begin{subfigure}{.45\textwidth}
  \centering
  \includegraphics[width=6cm,frame]{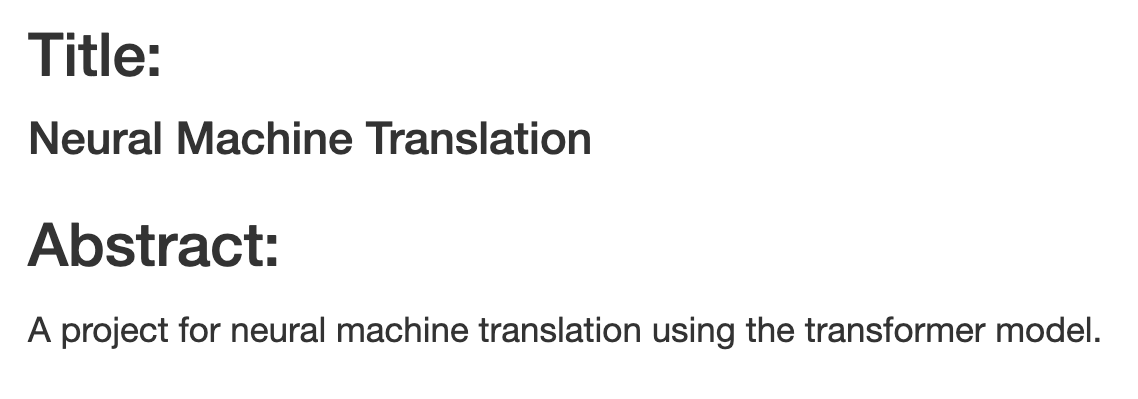}
  \caption{An example project proposal from the user about \texttt{neural machine translation}.}
  \label{fig:aan_proj}
\end{subfigure}%

\begin{subfigure}{.45\textwidth}
  \centering
  \includegraphics[width=6cm,frame]{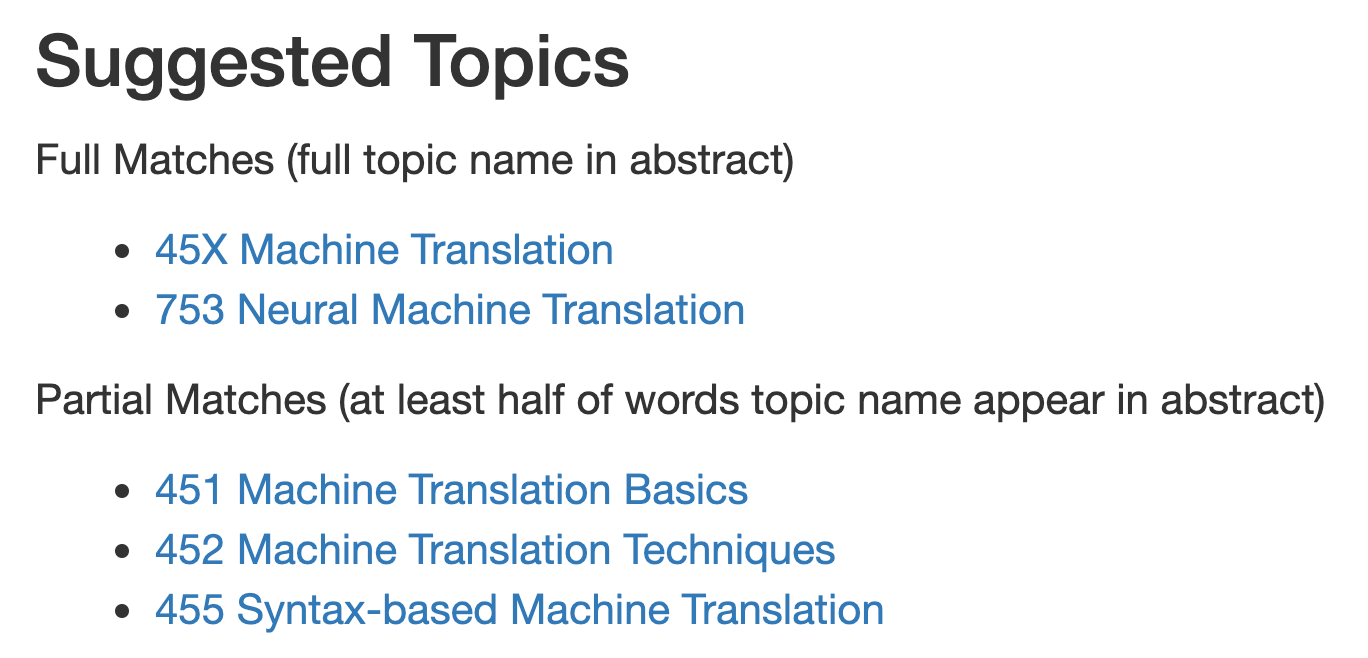}
  \caption{Suggested topics from the taxonomy. }
  \label{fig:aan_suggest_topic}
\end{subfigure}%

\begin{subfigure}{.45\textwidth}
  \centering
  \includegraphics[width=7cm,frame]{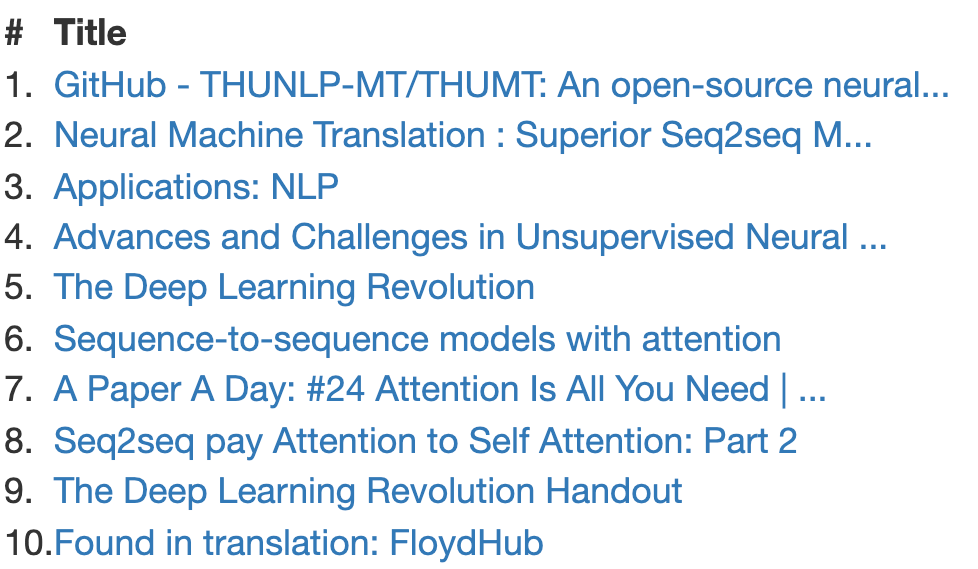}
  \caption{Suggested resources (the top 10 results).}
  \label{fig:aan_suggest_res}
\end{subfigure}
\caption{Resource Recommendation based on User Input.}
\label{fig:res_rec}
\end{figure}

Another interesting application is to make topic and resource recommendations based on a description of an actual project that a user wants to work on. We include an example in Figure \ref{fig:res_rec}.

The interface initially asks the user to input the title and a short abstract describing the project, as shown in Figure \ref{fig:aan_proj}. This sample project is about applying Transformers to do neural machine translation. Once the user submits this query, the system makes recommendations for both relevant concepts (Figure \ref{fig:aan_suggest_topic}) and possible resources to read (Figure \ref{fig:aan_suggest_res}). We can see that the suggested topics successfully capture the main query keyword neural machine translation. Besides, some suggested resources are also relevant for the query project, i.e., No. 2 is about using seq2seq models for neural machine translation. The search function, supported by the Apache Lucene Core library, sometimes misses important keywords in the abstract. In this case, the keyword (e.g., \texttt{transformer}) is ignored, and no relevant resources are suggested in the top list. In the future, we plan to improve this basic recommendation function by using better keyword extraction algorithms.

\subsection{Prerequisite Chain learning}
Prerequisite chain learning \cite{gordon-etal-2016-modeling} is used to help learners navigate through the space of topics within a domain by providing them with prerequisite concepts. A taxonomy like CLICKER can also help learn prerequisite chains for unknown concepts. This section discusses how to apply existing taxonomy relations to learn the prerequisite chain for unknown concepts in new domains. 

Existing work applies machine learning methods to solve this task by formulating it as a classification task \cite{gordon-etal-2016-modeling,li2019should,yu2020mooccube}: given a concept pair A and B, A$\to$B if A is a prerequisite concept of B. A typical method is to learn concept embeddings and conduct binary classification on the input (A, B): the label is positive if A$\to$B, negative otherwise. Materials used as learning concept relations include course content, video sequences, textbooks, lecture slides and Wikipedia articles \cite{pan-etal-2017-prerequisite,li2019should,yu2020mooccube}. LectureBankCD \cite{li2021unsupervised} is a dataset built for cross-domain prerequisite chain learning. It consists of labeled prerequisite concepts for different subjects, such as NLP (322 concepts) and Computer Vision (201 concepts). We then combine our taxonomy relations with their existing training set as the new training set. The evaluation is based on their test set. 

\textbf{Evaluation} We followed the work of \newcite{li2021unsupervised,li2021efficient} and first trained concept embeddings. Specifically, we built a Phrase2Vec (P2V) \cite{artetxe2018emnlp} embedding for each concept using the resources from the same dataset with \newcite{li2021unsupervised}.  Then, we compared three methods: logistic regression (\textbf{LR}), a single-layer neural network (\textbf{NN}) and a variational graph autoencoder \cite{li2019should} (\textbf{VGAE}). For each method, we compare our P2V embeddings (Pipeline+) and a basic pre-trained BERT model (BERT+)\footnote{\url{https://huggingface.co/bert-base-uncased}}. All classifiers are trained on NLP and then directly tested on CV. We show the results in Table \ref{tag:prereq_nlpcv}: in summary, the NN model performs the best, with a large improvement over the other two models, especially with respect to Accuracy. Moreover, applying the P2V embeddings trained on resources discovered by our pipeline quantitatively improves upon the BERT model in most cases when looking at Accuracy and the F1 score.

\begin{table}[t]
	\centering
	\small
		\begin{tabular}{@{}lcc@{}}\toprule
            \textbf{Model} & \textbf{Acc} & \textbf{F1}   \\ \midrule
            Logistic Regression & 0.5417 &	0.3654 \\
            VGAE \cite{li2019should}    & 0.5327	& 0.5869       \\
            2-layer Neural Network & \textbf{0.5868} &	\textbf{0.6024} \\
            
            
            \bottomrule 
            \end{tabular}
\caption{Transfer learning results for prerequisite chain prediction: NLP$\to$CV. We report Accuracy and F1 score.}
\label{tag:prereq_nlpcv}
\end{table}

\begin{figure}[t]
\centering
\begin{subfigure}{.42\textwidth}
  \centering
  \includegraphics[width=7cm]{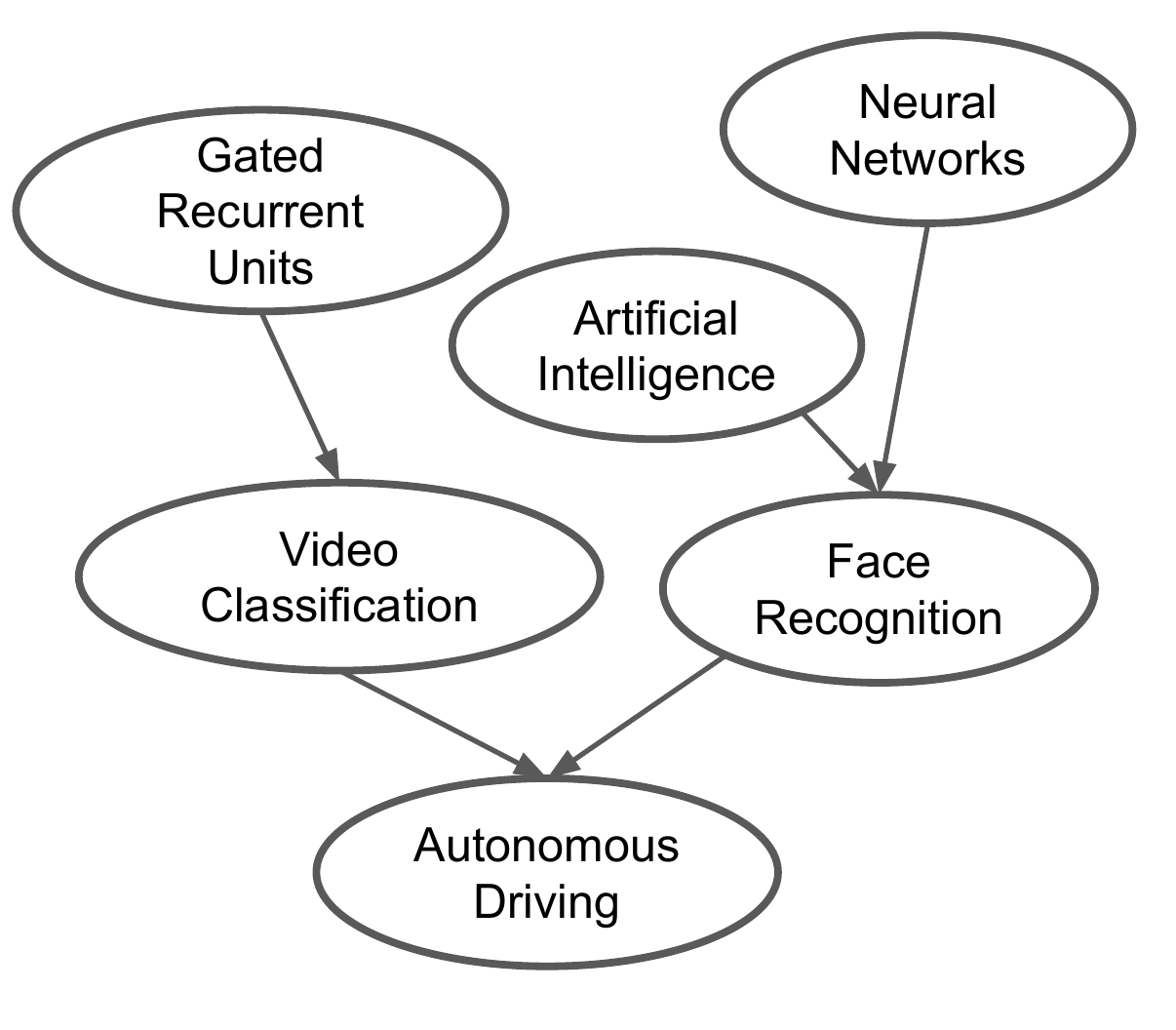}
  \caption{CV}
  \label{fig:prereq_cv}
\end{subfigure}%

\begin{subfigure}{.42\textwidth}
  \centering
  \includegraphics[width=7cm]{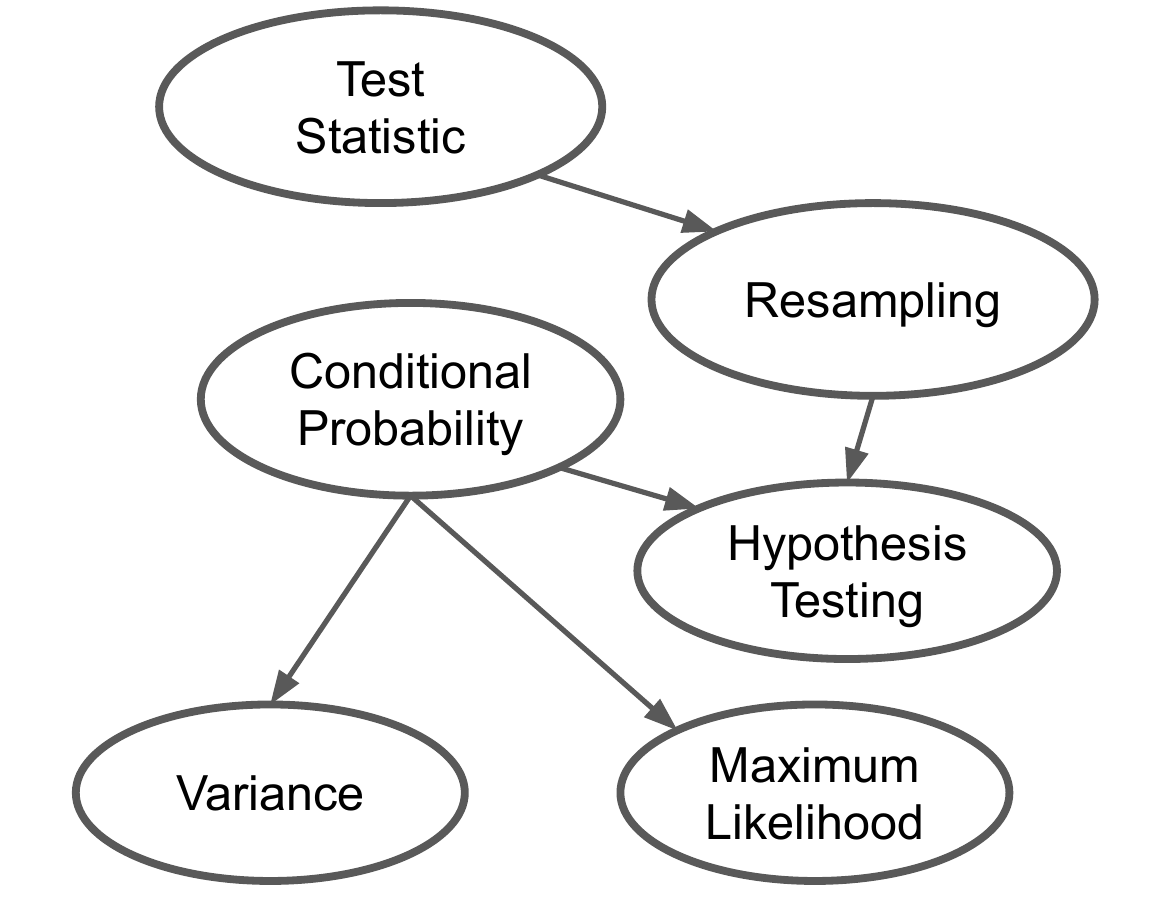}
  \caption{STATS}
  \label{fig:prereq_stats}
\end{subfigure}
\caption{Reconstructed concept graph from the best model.}
\label{fig:prereq}
\vspace{-3mm}
\end{figure}

\textbf{Case Study} We keep the best model, 2-layer Neural Network, and directly apply it on CV and STATS concepts respectively, in order to reconstruct the prerequisite concept graph. In Figure \ref{fig:prereq}, we show a portion of the concept graph from both domains. As can be seen, the model successfully captures correct relations, i.e., \textit{Video Classification} $\to$ \textit{Autonomous Driving} and \textit{Neural Networks} $\to$ \textit{Face Recognition} (Figure \ref{fig:prereq_cv}), \textit{Conditional Probability} $\to$ \textit{Variance} and \textit{Conditional Probability} $\to$ \textit{Maximum Likelihood} (Figure \ref{fig:prereq_stats}). However, some of the relation predictions have room for improvement. For instance, the connection from \textit{Artificial Intelligence} $\to$ \textit{Face Recognition} is overestimated, given that there are in reality several additional concepts in the path between the two.

\subsection{Survey Generation}
Obtaining new knowledge for a specific topic within the taxonomy can be important from a learner's perspective. To help readers get a quick understanding of this topic, a possible way is to do survey generation\cite{li2021surfer100}. People can get rid of searching multiple websites, textbooks, and other web resources to learn about a new topic. 
Survey generation aims to generate a survey for a query topic automatically \cite{deutsch2019summary}. Such a survey may contain a brief introduction, history, key ideas, variations, and applications in our scientific scenario. 

To achieve this, we followed the WikiSum \cite{liu2018generating} method to formulate this task as multi-document summarization. Given a query concept, we first search for relevant web pages and extract them as free texts using existing search engines like Google. For each section (i.e., history, main idea),  and pair the section name and extracted free texts as input to a pretrained BART \cite{lewis2019bart} model to generate a summary paragraph for each section.  Table \ref{tab:surveygen} shows the generated survey for the topic  \texttt{text summarization}. As an early attempt, one may notice some contents about real facts are not always accurate and coherent. For example, in the Key Ideas section, the model successfully captures the fact about extractive and abstractive summarization. But the last sentence talks about a new paper, which is irrelevant to the corresponding section. To the best of our knowledge, no other works attempt to apply neural methods to generate surveys using educational material, and our preliminary results show that this might be a very promising research direction and application. We leave improving survey quality as one of our future work.

\begin{table}[t]
\centering
\small
\begin{tabularx}{\columnwidth}{|X|}
\hline
\textbf{Introduction} \\ \hline
Text summarization is an interesting machine learning field that is increasingly gaining traction. As research in this area continues , we can expect to see breakthroughs that will assist in fluently and accurately shortening long text documents. In this article, we look at how machine learning can be used to help shorten text. \\
\hline
\textbf{History} \\ \hline
Summarization has been and continues to be a hot research topic in the data science arena. While text summarization algorithms have existed for a while , major advances in natural language processing and deep learning have been made in recent years. Google has reportedly worked on projects that attempt to understand novels. Summarization can help consumers quickly understand what a book is about.  \\
\hline

\hline
\textbf{Key Ideas} \\ \hline
Automatic summarization aims to produce a shorter version of an input text, preserving only the essential information. There are two main types of summarization : extractive summarization selects important sentences from the input and abstractive summarizing generates content without explicitly re-using whole sentences. In our new paper , we constructed two novel , large-scale summarization datasets from scientific journal articles.  \\

\hline
\textbf{Variations} \\ \hline
Multi-document summarization can be a powerful tool to quickly analyze dozens of search results. MeaningCloud 's Summarization API locates the most relevant phrases in a document and builds a synopsis with them. More specific summarization systems could be developed to analyze legal documents. \\
\hline

\hline
\textbf{Applications} \\ \hline
Summarization can be a crucial component in the tele-health supply chain when it comes to analyzing medical cases. The Spreading Activation approach does not allow to improve our results. Tables 8 and 9 show the high recall obtained with these methods, which may be a very interesting feature in some cases. \\
\hline
\end{tabularx}
\caption{Sample survey generation of the topic \texttt{Text Summarization}. }
\label{tab:surveygen}
\vspace{-3mm}
\end{table}

\section{Conclusion}
In this work, we introduced CLICKER, a practical CL/NLP classification scheme for educational resources. We also showed how CLICKER can make a difference in five applications: educational platform, resource retrieval, resource recommendation, prerequisite chain learning, and survey generation.

\bibliography{anthology,custom}
\bibliographystyle{acl_natbib}

\end{document}